\documentclass[runningheads]{llncs}
\usepackage[T1]{fontenc}
\usepackage{cite}
\usepackage{amsmath,amssymb,amsfonts}
\usepackage{algpseudocode}
\usepackage{graphicx}
\usepackage{textcomp}
\usepackage{xcolor}
\usepackage{float}
\usepackage{caption}
\usepackage{url}
\usepackage{multirow}
\usepackage{hyperref}
\usepackage{svg}
\usepackage{algorithm}
\usepackage{array}

\usepackage{graphicx}
\usepackage{amsmath}

\begin{document}

\title{Markov-Enhanced Clustering for Long Document Summarization: Tackling the 'Lost in the Middle' Challenge with Large Language Models}

\titlerunning{Markov-Enhanced Clustering for Long Document Summarization}


\author{Aziz Amari\inst{1}\orcidID{0009-0005-7020-3051}\and Mohamed Achref Ben Ammar\inst{1}\orcidID{0009-0007-9741-2405} 
}

\authorrunning{A. Amari and M. Ben Ammar}
\institute{National Institute of Applied Science and Technology (INSAT), University of Carthage, Tunis, Tunisia \\ 
\email{aziz.amari@insat.ucar.tn}, \email{mohamedachref.benammar@insat.ucar.tn}} 

\maketitle              % typeset the header of the contribution

\begin{abstract}
The rapid expansion of information from diverse sources has heightened the need for effective automatic text summarization, which condenses documents into shorter, coherent texts. Summarization methods generally fall into two categories: extractive, which selects key segments from the original text, and abstractive, which generates summaries by rephrasing the content coherently. Large language models have advanced the field of abstractive summarization, but they are resource-intensive and face significant challenges in retaining key information across lengthy documents, which we call being "lost in the middle". To address these issues, we propose a hybrid summarization approach that combines extractive and abstractive techniques. Our method splits the document into smaller text chunks, clusters their vector embeddings, generates a summary for each cluster that represents a key idea in the document, and constructs the final summary by relying on a Markov chain graph when selecting the semantic order of ideas.

\end{abstract}

\section{Introduction}
In recent years, the growing volume of information from diverse sources has heightened the need for effective automatic text summarization \cite{atssurvey}. Applications of summarization span across various fields, including information retrieval, content creation, and extracting insights from extensive datasets.

Summarization is defined as the process of creating shorter, coherent texts that accurately reflect the original content of a document \cite{sumtech}. This process is generally approached in two main ways: extractive and abstractive \cite{Pang2022}.

Abstractive summarization generates a summary by interpreting and rephrasing the original content, aiming for semantic coherence and linguistic fluency. This method mirrors human summarization techniques more closely, as it involves understanding and rewording ideas rather than merely copying text. However, abstractive models can sometimes lose meaning or omit important ideas, as the process of rephrasing may lead to oversimplification or the exclusion of critical information from the original document.

In contrast, extractive summarization involves selecting and assembling key segments—such as words, phrases, or sentences—directly from the source document. While it efficiently captures the most important parts of the text, this approach does not take semantic coherence or fluency into consideration.

Although traditional natural language processing (NLP) techniques \cite{rahul2020} have been effective, they often face challenges when dealing with long documents. This difficulty can lead to the loss of key information and a lack of coherence, especially in summarizing extensive texts like business reports, video transcriptions, and books \cite{Koh2022}.

Transformer-based methods, including those based on BERT \cite{devlin-etal-2019-bert}, offer significant improvements in summarization by leveraging self-attention mechanisms to understand context and relationships between words, resulting in more coherent summaries \cite{Miller2019}. However, despite these advancements, such models still struggle with the memory and computational requirements needed for processing very large or complex documents.

Large language models (LLMs) \cite{Vaswani2017} have excelled, particularly in abstractive summarization, and are driving significant advancements in this area \cite{Zhang2024}. Their ability to produce summaries that are semantically rich and contextually relevant marks a substantial leap forward \cite{tam-etal-2023-evaluating}. However, these models are resource-intensive and may encounter issues like "lost in the middle" \cite{liu-etal-2024-lost}, particularly in maintaining context across lengthy documents.

To address these challenges, we propose a hybrid summarization method that combines extractive and abstractive techniques to better handle large documents. Our method splits the document into smaller text chunks, clusters their vector embeddings, generates a summary for each cluster that represents a key idea in the document, and constructs the final summary by relying on a Markov chain graph when selecting the semantic order of ideas. Technical details will be explored in the next sections.

The structure of this paper is as follows: Section 2 introduces the dataset, Section 3 covers the methodology, Section 4 presents and discusses the results, and Section 5 concludes the paper.

\section{Materials}
For our research, we initially explored multiple long-document summarization datasets, including SCROLLS \cite{shahamscroll} and FacetSum \cite{mengsum}, but ultimately selected BookSum \cite{Kryscinski2021} as our primary dataset due to its superior document length characteristics. While SCROLLS and FacetSum offer valuable document collections, their average document lengths (approximately 8.4k and 7.3k tokens, respectively) are substantially shorter than BookSum’s average of 108k tokens per document.
The extensive length of BookSum documents makes the dataset particularly well-suited for evaluating the effectiveness of our proposed method in true long-document scenarios. Since our method involves chunking the document and clustering its vector embeddings (detailed in the following section), short documents yield a very limited number of vectors, resulting in less meaningful clustering and ultimately degraded performance. In contrast, long documents provide a richer structure for clustering, which is where our method excels.
With 187 books paired with summaries written by humans, BookSum closely mirrors long-form content in the real world such as business reports, books, and extended transcriptions, offering a more challenging and realistic test environment, making it the optimal choice for our study.

\begin{figure}
    \centering
  \includegraphics[width=0.7\textwidth]{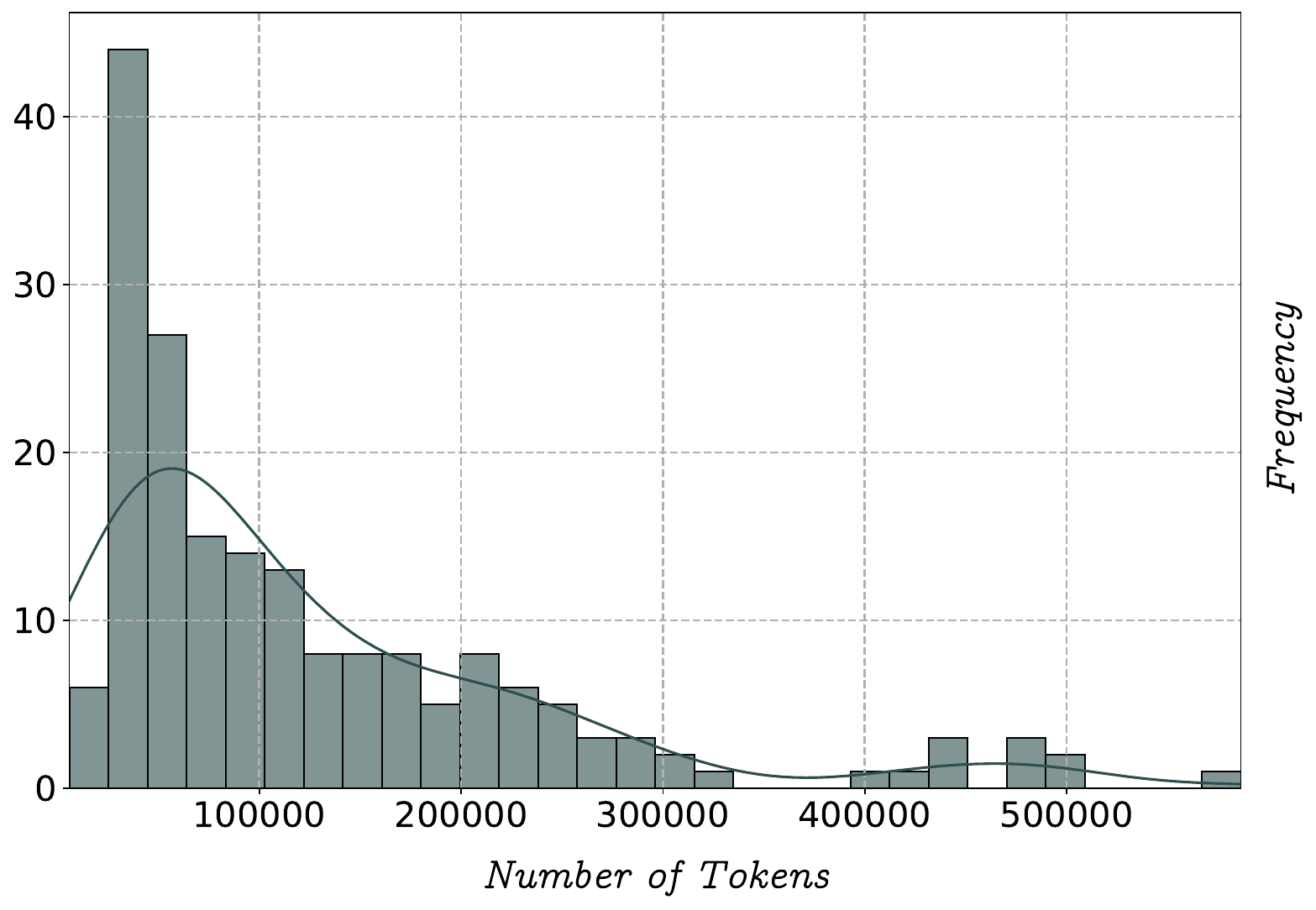}
      \captionsetup{justification=centering}

  \caption{Variability of Book Length in Booksum, Measured by Token Count}
  \label{fig:tokens}
\end{figure}

Figure~\ref{fig:tokens} illustrates the distribution of book lengths in terms of the number of tokens in the selected dataset.
Further analysis of the dataset reveals an average compression ratio of 2\%, meaning that the length of the generated summaries is, on average, 1/50 of the length of the original documents. This compression ratio reflects the extent to which the summaries condense the original content while retaining essential information.

\begin{figure}[h!]
  \centering
  \includegraphics[width=0.7\textwidth]{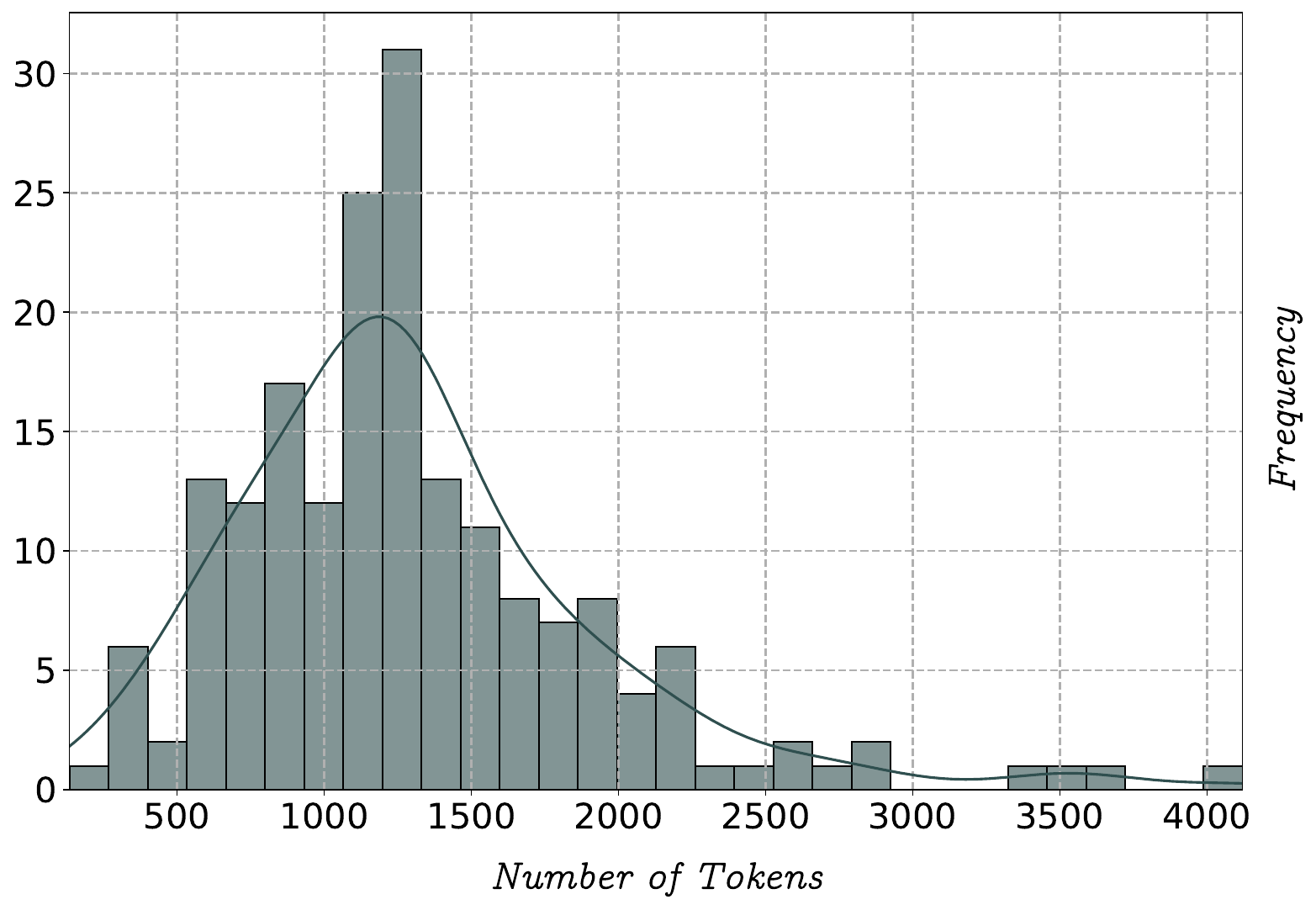}
    \captionsetup{justification=centering}

  \caption{Variability of Summary Length in the Dataset, Measured by Token Count}
  \label{fig:compression}
\end{figure}

Figure~\ref{fig:compression} presents the distribution of summary lengths across the dataset, with an average summary length of 1,301 tokens.

\section{Methods}
\subsection{Our Approach}

\begin{figure}[t!]
  \includegraphics[width=\textwidth]{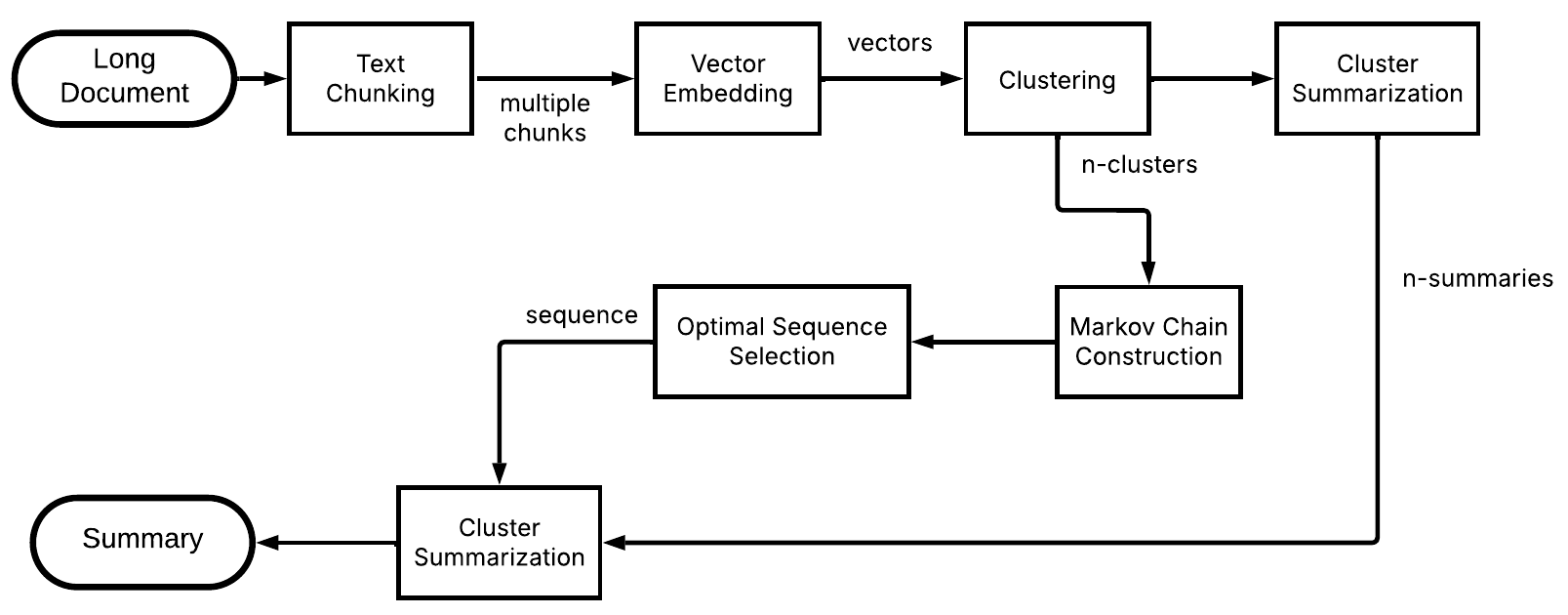}
  \caption{Pipeline illustrating our proposed summarization method}
  \label{fig:approach}
\end{figure}

Figure~\ref{fig:approach} provides an overview of our summarization pipeline. The process begins with splitting the document into chunks. Each chunk is represented as a numerical vector using a text embedding model. These vectors are then clustered to identify key ideas within the document. Summaries are created for each cluster. The optimal sequence of summaries is selected after constructing a Markov chain that models the semantic order of clusters. Finally, an LLM aggregates them and outputs the final summary.
\subsubsection{Text Chunking}
Text chunking \cite{Finardi2024} is used to divide long documents into smaller, manageable sections. This method involves splitting the document into chunks with a length of 500 tokens and a 20-token overlap between consecutive chunks. The overlap is essential for retaining links between chunks and preserving the necessary context. Although recursive chunking methods can be faster and simpler, using text chunking with overlap ensures continuity and coherence throughout the document.

\subsubsection{Vector Embedding Generation}
Following the chunking process, each extracted chunk is transformed into a high-dimensional vector. Unlike large language models (LLMs), embedding models \cite{Patil2023} are less resource-intensive and are specifically designed to capture semantic meaning. In natural language processing (NLP) applications, vector embeddings are used to convert textual data into a numerical format that preserves contextual relationships within the text, enabling efficient analysis, search, and manipulation.

We use the nomic-embed-text-v1 embedding model \cite{Nussbaum2024}, a fully reproducible, open-source model with open weights and open data, supporting an 8192 context length for English text. The generated vectors form the basis for subsequent clustering, as they encode the semantic content of the text in a way that facilitates the grouping of similar chunks. This method ensures that the resulting clusters represent thematic groups.

\subsubsection{Clustering and Cluster Summarization}

After generating vector embeddings from the document chunks, we use Kmeans++ clustering \cite{Arthur2007}  to organize these vectors into meaningful thematic groups. This method groups similar vectors, revealing the underlying themes within the document.

Following the clustering process, we proceed with cluster summarization. This involves selecting the $top\_k$ vectors, with \( k=5 \), closest to the centroid of each group and summarizing the original text of these representative chunks using GPT-4o-mini \cite{openai_gpt4o_mini}. We selected GPT-4o Mini because our primary objective is to assess whether summarization improves with our method rather than to compare different LLMs. This ensures that our analysis focuses on the effectiveness of our approach rather than variations in model performance.

\subsubsection{Markov Chain Modeling}
Then, we construct a Markov chain to model the transition probabilities between clusters in a document. A Markov chain is a mathematical system that undergoes transitions from one state to another according to certain probabilistic rules. In our case, the states correspond to different clusters, and the transitions between these states are modeled based on their observed frequencies.
Let \( S = \{s_1, s_2, \dots, s_n\} \) be the sequence of cluster IDs corresponding to the \( n \) chunks in their sequential order from the document. To construct the transition matrix \( T \in \mathbb{R}^{k \times k} \), where \( k \) is the number of unique clusters, we define:

\begin{equation}
    T_{i,j} = \frac{\text{count}(s_{t} = i, s_{t+1} = j)}{\sum_{j=1}^{k} \text{count}(s_{t} = i, s_{t+1} = j)}
\end{equation}

Here, \( \text{count}(s_{t} = i, s_{t+1} = j) \) represents the number of transitions from cluster \( i \) to cluster \( j \) observed in the sequence \( S \). Each row of the matrix \( T \) sums up to 1, representing a probability distribution over the next possible clusters given the current cluster. Thus, \( T_{i,j} \) denotes the probability of transitioning from cluster \( i \) to cluster \( j \).

This matrix \( T \) captures the transition probabilities between clusters, treating the cluster IDs as states, and considering their sequential order in the document. By modeling these transitions, the Markov chain reflects the narrative flow of the text, showing how thematic elements progress and relate to one another in the original document.
\subsubsection{Pathfinding in Transition Matrix}
In the previous step, we constructed the transition matrix \( T \) for our Markov chain. Using \( T \), we analyze the document's narrative structure by identifying the most probable Hamiltonian path \cite{Uehara2005}. This path connects all nodes without revisiting any, visits each node exactly once, maximizes the product of transition probabilities between consecutive nodes, and does not require specifying starting or ending nodes. This problem is NP-hard \cite{Hochba1997} as it is a variant of the Hamiltonian path problem.

To solve this efficiently, we employ a dynamic programming (DP) approach with bitmasking. Let \( \text{dp}(S, i) \) represent the maximum probability of a path ending at node \( i \), where \( S \) is a subset of nodes encoded as a bitmask. The recurrence relation updates \( \text{dp}(S, i) \) by iterating over all nodes \( j \) in \( S \setminus \{i\} \), computing \( \text{dp}(S \setminus \{i\}, j) \cdot T[j][i] \), and storing the predecessor of \( i \) in a parent table for path reconstruction. The algorithm iterates over all subsets \( S \) (from size 1 to \( n \)), achieving a time complexity of \( O(n^2 \cdot 2^n) \) (see Appendix~\ref{app:algorithm} for pseudocode).

Figure~\ref{fig:example_graph} illustrates an example graph with \( n = 3 \) clusters and the optimal Hamiltonian path \( 1 \to 3 \to 2 \). This example graph demonstrates how the transition matrix \( T \) is used to identify the most probable Hamiltonian path in a small-scale scenario.
\begin{figure}[b]
  \centering
  \includegraphics[width=0.5\textwidth]{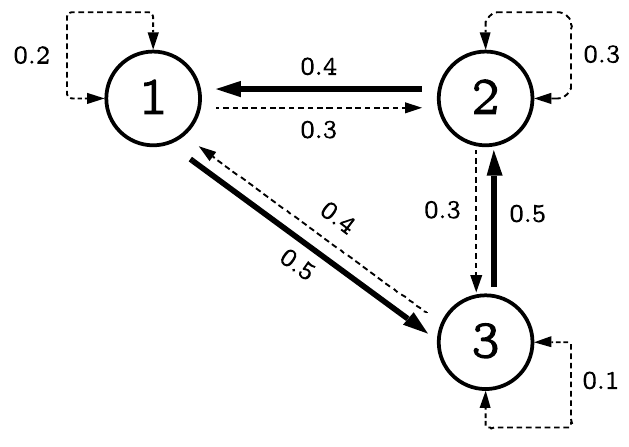}
  \captionsetup{justification=centering}
  \caption{Example graph with \( n = 3 \) clusters. Solid lines indicate the optimal Hamiltonian path.}
  \label{fig:example_graph}
\end{figure}

\begin{table*}[h!]
  \centering
  \renewcommand{\arraystretch}{1.4}
  
  \setlength{\tabcolsep}{15pt}
  \begin{tabular}{||l|c|c||}
    \hline
    \textbf{Method} & \textbf{Time} & \textbf{Space} \\
    \hline
    \hline
    \textbf{Brute Force} & \( O(n \cdot n!) \) & \textbf{\( O(n) \)} \\    
    \hline
    \textbf{DP} & \textbf{\( O(n^2 \cdot 2^n) \)} & \( O(n \cdot 2^n) \) \\
    \hline
  \end{tabular}
  \vspace{8pt}
  \captionsetup{justification=centering}
  \caption{Complexity Comparison for Finding the Most Probable Hamiltonian Path}
  \label{tab:complexities}
\end{table*}

The brute force approach, which evaluates all node permutations, becomes impractical when \( n \) reaches around 11 due to its slow performance. Instead, we use a dynamic programming (DP) approach, which reasonably handles graphs with up to 22 nodes, providing solutions in a few seconds on a typical host machine. The complexity analysis of these methods is summarized in Table~\ref{tab:complexities}.

\subsubsection{Final Summary Generation}
To obtain the final abstractive summary, we concatenate the summaries of each cluster in the sequence determined by the Hamiltonian path and feed the concatenated summaries to an LLM. This approach ensures that the final summary is coherent and captures the essential information from the document.

\subsection{Other Approaches}
Two alternative methods for document summarization are considered for comparison against our approach. The first method is a baseline approach where an LLM generates a summary directly from the entire document. The second method uses a cluster-based approach \cite{Keswani2024} but does not include the sequence selection step for analyzing relationships between clusters using a Markov chain. Results obtained using all these approaches will be compared in Section 4.

\subsection{Evaluation}
We evaluate the quality of summaries using a combination of metrics: ROUGE \cite{lin2004rouge}, BERTScore \cite{zhang2019bertscore}, coherence \cite{steen2022coherence} metrics, and BLEURT \cite{Sellam2020} to ensure a comprehensive assessment.

ROUGE (Recall-Oriented Understudy for Gisting Evaluation) is a metric commonly used to evaluate the quality of extractive summarization by measuring the overlap between the candidate summary and reference summaries.  This evaluation provides insights into the completeness and relevance of the summary, ensuring that it covers the necessary information and retains critical details from the source material.

Coherence assesses the logical flow and thematic consistency of the text. First-order coherence measures the cosine similarity between consecutive sentences to gauge local coherence, while second-order coherence evaluates the similarity between sentences that are two sentences apart, offering insights into broader thematic consistency.

BERTScore evaluates semantic similarity by using contextual embeddings from BERT. Measures precision, recall, and F1 score based on token embeddings, providing a detailed analysis of semantic meaning and contextual relevance.

BLEURT is an evaluation metric for Natural Language Generation. It takes a pair of sentences as input, a reference and a candidate, and it returns a score that indicates to what extent the candidate is fluent and conveys the meaning of the reference. It is comparable to sentence-BLEU, BERTscore, and COMET.

Together, these metrics ensure that summaries are coherent and semantically accurate.

\section{Results and Discussion}
Table~\ref{tab:evaluation_results} presents summarization results using ROUGE, BERTScore, BleuRT, and coherence metrics. GPT-4o Mini \cite{openai_gpt4o_mini} was used in all implementations.

\begin{table}[h!]
\centering
\renewcommand{\arraystretch}{1.2} % Increase row spacing
\setlength{\tabcolsep}{4pt}      % Adjust column separation
\begin{tabular}{||l||>{\centering\arraybackslash}p{1.2cm} >{\centering\arraybackslash}p{1.2cm} ||
>{\centering\arraybackslash}p{1.2cm} >{\centering\arraybackslash}p{1.2cm} || 
>{\centering\arraybackslash}p{1.2cm} >{\centering\arraybackslash}p{1.2cm}||}
\hline\hline
% First row: Dataset label (BookSum)
 & \multicolumn{6}{c||}{\textbf{Evaluation Metrics}} \\
\cline{2-7}

% Second row: Grouping of metrics
\textbf{Approach}  & 
\multicolumn{2}{c||}{\textbf{ROUGE}} &  
\multicolumn{2}{c||}{\textbf{Coherence }} &  
\multicolumn{2}{c||}{\textbf{Semantic }}  \\
\cline{2-7}

% Third row: Individual metric headers
 & \textbf{R-1} 
 & \textbf{R-2} 
 & \textbf{1st-O} 
 & \textbf{2nd-O}  
 & \textbf{BF1}  
 & \textbf{BLRT} 
\\
\hline\hline

% Rows for each approach

\textbf{Cluster-Sum}                
 & 33.72 & 6.190 & 0.852 & 0.851 & 0.825 & 0.827  \\
 \hline
\textbf{LLM-Full}           
 & 23.99 & 5.440 & 0.729 & 0.731 & \textbf{0.831} & 0.826 \\
 \hline
 \textbf{Markov-Cluster}  
 & \textbf{34.13} & \textbf{6.392} & \textbf{0.863} & \textbf{0.862} & 0.823 & \textbf{0.829} \\
\hline\hline

\end{tabular}
\vspace{3pt} % Adjust the space as needed
      \captionsetup{justification=centering}
\captionsetup{justification=centering}
\caption{Evaluation results on the BookSum dataset. 
R-1: ROUGE-1, R-2: ROUGE-2, 1st-O: First-Order Coherence, 2nd-O: Second-Order Coherence, BRT-F1: BertF1 Score, BLRT: BleuRT Score
}
\label{tab:evaluation_results}
\end{table}

The results demonstrate that our proposed approach achieves the highest ROUGE-1 and ROUGE-2 scores, indicating that it effectively balances extractiveness while retaining key information compared to directly using the LLM on the full document. This superior performance suggests that our method captures essential elements of the text more effectively.

Furthermore, our approach achieves the best scores in both 1st and 2nd order coherence, along with the highest BleuRT score, highlighting its ability to generate summaries with a logical and semantically consistent flow of ideas while preserving semantic fidelity. This can be attributed to the semantic linking process introduced by the Markov Chain model, which selects the most coherent sequence of ideas, as opposed to simply summarizing chunks of text in isolation.

While GPT-4o Mini’s training dataset remains closed-source, the possibility that BookSum was included in its pretraining corpus does not impact our evaluation. Our results demonstrate that simply feeding the full document to the LLM is less effective than applying our approach. The improvement seen in ROUGE, coherence, and BleuRT metrics confirms that our clustering and Markov-based linking contribute directly to better summarization quality, rather than any prior knowledge the LLM may have of the dataset.

Overall, these findings affirm that our hybrid approach successfully integrates extractive and abstractive summarization, leading to significant improvements across all evaluated metrics.

\section{Conclusion}
In summarizing large documents, our proposed approach effectively preserves key information and improves summarization metrics, addressing the issue of getting "lost in the middle" that often occurs with direct LLM-based summarization. By integrating both extractive and abstractive techniques, our hybrid method successfully manages the complexities of lengthy texts, capturing their essence while maintaining coherence in the generated summaries. Notably, our approach is significantly less resource-intensive compared to direct LLM inference, as the process of generating embeddings, clustering, and applying the Markov Chain model is computationally more efficient than feeding the entire document into an LLM. This efficiency is achieved by processing only a small subset of the data within the LLM’s context window, rather than handling the entire document at once.  

Looking ahead, we plan to enhance our approach by focusing on several key areas. We aim to develop improved evaluation metrics to better assess summarization effectiveness. Additionally, we will explore a greedy sequence selection strategy to efficiently handle large values of \( n \). Benchmarking with various LLMs, embedding models, and clustering techniques will be conducted to identify the most effective configurations. Furthermore, we intend to extend our method to diverse datasets beyond books, including different types of long documents such as business reports and technical papers.  

A crucial aspect requiring further evaluation is determining whether limiting clustering to values of \( n < 22 \) improves performance or negatively impacts summarization quality Conducting a thorough analysis will help assess whether this constraint optimally balances computational efficiency and information retention or if larger cluster sizes remain necessary for capturing key content effectively. The code for this paper is available on GitHub\footnote{Code Repository : \\ \url{https://github.com/achrefbenammar404/large_doc_summarization.git} }.

\newpage
\appendix
\section{Most Probable Hamiltonian Path Algorithm} \label{app:algorithm}

\begin{algorithm}[H]
\caption{Most Probable Hamiltonian Path in a Markov Chain} \label{alg:hamiltonian}
\textbf{Input:} Transition matrix \( T \) of size \( n \times n \). \\
\textbf{Output:} Maximum probability \( \text{max\_prob} \), path \( \text{path} \).
\begin{algorithmic}[1]
\State Initialize DP table \( \text{dp} \) and \( \text{parent} \) tables
\For{each node \( i \in \{0, 1, \dots, n-1\} \)}
    \State \( \text{dp}[1 \ll i][i] \gets 1 \) \Comment{Base case: start at node \( i \)}
\EndFor
\For{each subset \( S \in [1, 2^n - 1] \)} \Comment{Iterate in order of increasing \( |S| \)}
    \For{each node \( i \in S \)} \Comment{Possible end node of the path}
        \For{each node \( j \in S \setminus \{i\} \)}
            \If{\( \text{dp}[S \setminus \{i\}][j] \cdot T[j][i] > \text{dp}[S][i] \)}
                \State \( \text{dp}[S][i] \gets \text{dp}[S \setminus \{i\}][j] \cdot T[j][i] \)
                \State \( \text{parent}[S][i] \gets j \) \Comment{Track predecessor}
            \EndIf
        \EndFor
    \EndFor
\EndFor
\State \( \text{max\_prob} \gets \max_{i} \text{dp}[2^n - 1][i] \) \Comment{Find maximum probability}
\State \( \text{end\_node} \gets \arg\max_{i} \text{dp}[2^n - 1][i] \)
\State Reconstruct \( \text{path} \) by backtracking from \( \text{end\_node} \) using \( \text{parent} \)
\State Reverse \( \text{path} \) to get the start-to-end order
\end{algorithmic}
\end{algorithm}

\textbf{Complexity:} The algorithm runs in \( O(n^2 \cdot 2^n) \) time and \( O(n \cdot 2^n) \) space complexities, as analyzed in Table~\ref{tab:complexities}.

\end{document}